\documentclass[journal]{IEEEtran}

\usepackage{amsfonts,amssymb,amsmath}
\usepackage{bm}

\usepackage{algorithm}
\usepackage{algorithmic}
\usepackage{cite}
\usepackage{makecell}
\usepackage{graphicx}
\usepackage{textcomp}
\usepackage{multirow}
\usepackage{booktabs}
\usepackage{float}
\usepackage{stfloats}
\usepackage{url}
\usepackage{graphics}
\usepackage{subfigure}
\usepackage{colortbl}
\usepackage{setspace}
\usepackage{array}
\usepackage{soul}
\soulregister\cite7
\soulregister{\ref}7
\ifCLASSINFOpdf
\else
\fi
\begin{document}
%
\title{Feature Encoding with AutoEncoders for Weakly-supervised Anomaly Detection} 
%
%

\author{Yingjie~Zhou,~\IEEEmembership{Member,~IEEE,}
        Xucheng~Song,
        Yanru~Zhang,~\IEEEmembership{Member,~IEEE,}
        Fanxing~Liu,
        \\Ce~Zhu,~\IEEEmembership{Fellow,~IEEE,}
        and~Lingqiao~Liu,~\IEEEmembership{Member,~IEEE}
\thanks{Submitted to IEEE Transactions on Neural Networks and Learning Systems Special Issue on ``Deep Learning for Anomaly Detection". This work is partly supported by National Natural Science Foundation of China (NSFC) with grant number 61801315. \emph{(Corresponding author: Lingqiao Liu)}}
\thanks{Yingjie Zhou is with the College of Computer Science, Sichuan University (SCU), China (email: yjzhou@scu.edu.cn; yjzhou09@gmail.com).}
\thanks{Xucheng Song is with the School of Information and Communication Engineering, University of Electronic Science and Technology of China (UESTC), China. He is also with the College of Computer Science, Sichuan University (SCU), China (email: xucheng\_song@foxmail.com).}
\thanks{Yanru Zhang is with the School of Computer Science and Engineering, University of Electronic Science and Technology of China (UESTC), China (email: yanruzhang@uestc.edu.cn).}
\thanks{Fanxing Liu is with the College of Software Engineering, Sichuan University (SCU), China (email: finwarrah@gmail.com).}
\thanks{Ce Zhu is with the School of Information and Communication Engineering, University of Electronic Science and Technology of China (UESTC), China (email: eczhu@uestc.edu.cn).}
\thanks{Lingqiao Liu is with the School of Computer Science, University of Adelaide (email: lingqiao.liu@adelaide.edu.au).}
}

%
%

\markboth{the manuscript has been published by IEEE (IEEE TNNLS, 2021, DOI: 10.1109/TNNLS.2021.3086137). Please do not cite the preprint}%
{Shell \MakeLowercase{\textit{et al.}}: Bare Demo of IEEEtran.cls for IEEE Journals}
%



\maketitle

\begin{abstract}
Weakly-supervised anomaly detection aims at learning an anomaly detector from a limited amount of labeled data and abundant unlabeled data.  Recent works build deep neural networks for anomaly detection by discriminatively mapping the normal samples and abnormal samples to different regions in the feature space or fitting different distributions. However, due to the limited number of annotated anomaly samples, directly training networks with the discriminative loss may not be sufficient. To overcome this issue, this paper proposes a novel strategy to transform the input data into a more meaningful representation that could be used for anomaly detection. Specifically, we leverage an autoencoder to encode the input data and utilize three factors, hidden representation, reconstruction residual vector, and reconstruction error, as the new representation for the input data. This representation amounts to encode a test sample with its projection on the training data manifold,  its direction to its projection and its distance to its projection. In addition to this encoding, we also propose a novel network architecture to seamlessly incorporate those three factors. From our extensive experiments, the benefits of the proposed strategy are clearly demonstrated by its superior performance over the competitive methods. Code is available at: https://github.com/yj-zhou/Feature\_Encoding\_with\_AutoEncoders\_for\_Weakly-superv\\ised\_Anomaly\_Detection.
\end{abstract}


%
\IEEEpeerreviewmaketitle

\section{Introduction}
%
%
%
%
\IEEEPARstart{A}{nomaly} detection is an important and fundamental task for broad domains, including networking\cite{jiang2018deep},\cite{nicolau2018learning},\cite{miao2018distributed},\cite{xie2018line}, machine learning\cite{chen2019unsupervised},\cite{xu2018unsupervised},\cite{zong2018deep},\cite{zhang2018anomaly}, and computer vision\cite{luo2019video},\cite{leyva2017video},\cite{pang2020self}. It has been well investigated in a number of application scenarios, and many algorithms have excellent performance on related public datasets. However, it is still challenging to detect anomalies when there is no or only a few annotated anomaly samples, in spite of that abundant normal data can be easily accessible. This is commonly seen in practice, \emph{e.g.}, network anomaly detection\cite{nicolau2018learning}, finance fraud detection\cite{zhou2019model} and video surveillance\cite{luo2019video}. 

Depending on whether the anomaly samples are available at the training stage, the existing anomaly detection approaches can be categorized into three groups, \emph{i.e.}, supervised learning methods\cite{jiang2018deep},\cite{yang2018active},\cite{liu2021deeppayload}, unsupervised learning methods\cite{nicolau2018learning},\cite{zong2018deep},\cite{8794857},\cite{ergen2019unsupervised},\cite{pang2018learning}, \cite{dong2019mem} and weakly-supervised learning methods\cite{pang2019deep},\cite{ruff2019deep},\cite{pang2019deepweakly}. 
The supervised methods leverage both normal and abnormal data to train a detector. In order to achieve satisfactory performance, both data categories are usually required to be relatively balanced. 
The unsupervised methods attempt to build a generative model for modeling normal data, which could detect abnormal samples as outliers to the generative model. 
The weakly-supervised methods address the scenario in which a few annotated anomaly samples are provided at the training stage and learn an anomaly detector from the imbalance training data.
These approaches make use of the limited number of anomalies by fitting the normal samples and abnormal samples to different distributions. 
However, learning such an anomaly detector can be challenging since the model tends to over-fit the limited amount of abnormal data. Recently, deep learning based anomaly detection\cite{pang2019deep},\cite{ruff2019deep},\cite{ruff2018deep},\cite{pang2020deep},\cite{10.1145/3219819.3220024} has become increasingly popular due to its capability of learning feature representations. For the deep learning based anomaly detector, the over-fitting issue could be in particular severe due to its relatively larger number of parameters.

\begin{figure}[!h]
\centering
\includegraphics[width=5.6cm,height=6.4cm]{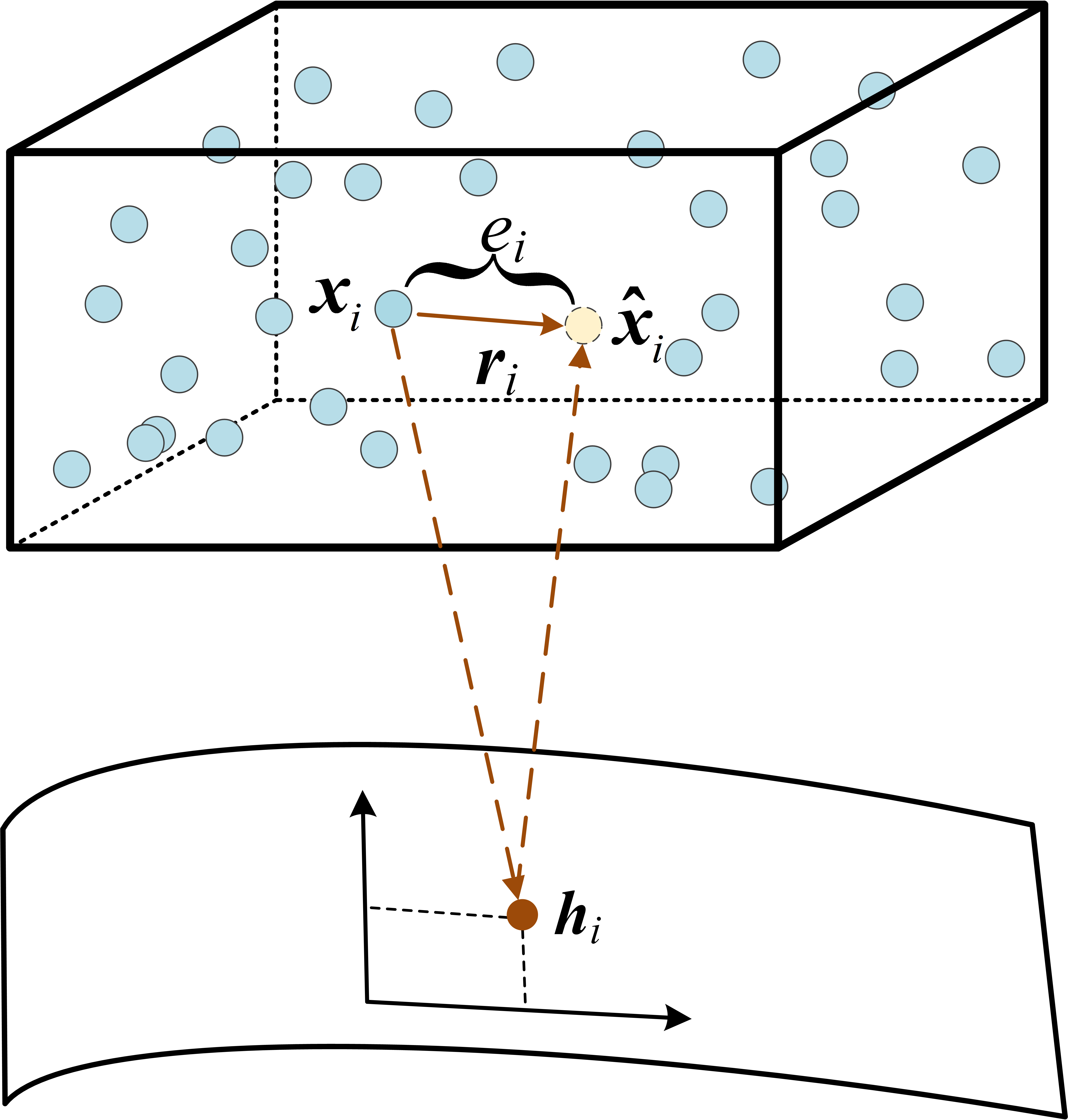}
\centering
\caption{An illustration of the proposed encoding strategy. An input sample
$\bm{x}_i$ is transformed into a feature representation with three factors \(\bm{h}_{i}\), \(\bm{r}_{i}\) and \({e}_{i}\), which can be approximately seen as the sample’s intrinsic coordinate of its projection on the training data manifold \(\bm{h}_{i}\), the direction between the projection and the original data point \(\bm{r}_{i}\) and the distance \({e}_{i}\), respectively. }
\label{manifoldspace}
\end{figure}

To overcome this issue, this paper proposes a deep weakly-supervised anomaly detection method that marries the methodologies from unsupervised anomaly detection with semi-supervised anomaly detection. The key idea is to leverage the autoencoder, a commonly used unsupervised anomaly detector, to fit the normal data and then extract a feature representation from that for weakly-supervised anomaly detection. The rationale is that the outlier could become more distinguishable by using the representation derived from the autoencoder which somehow uncovers some prior knowledge for the normal samples. 
Specifically, the proposed feature representation is the concatenation of three factors derived from an autoencoder, \emph{i.e.}, hidden representation, reconstruction residual vector, and reconstruction error, which are extracted as the feature presentation for each input sample. The design of such a representation also has a geometric implication. As shown in Figure \ref{manifoldspace}, the autoencoder essentially transforms the (normal) data manifold into a lower-dimensional space, in which the hidden representation, \emph{i.e.}, the first factor, can be viewed as the intrinsic space coordinates for the original high dimensional data. The second factor, \emph{i.e.}, the reconstruction residual vector, can be seen as the offset from an input sample to its projection on the normal data manifold. The last factor encodes the distance between the input sample and its projection on the manifold.
A typical anomaly will significantly deviate from the normal samples in the view of the feature representation. The deviation could be manifested in one of the following cases: unusual intrinsic coordinates, abnormal offset, large distance to the normal data manifold, or any possible combinations of the former three.
Once the feature representation is extracted, it will be fed into a specially designed subnetwork for anomaly detection. The subnetwork is modified from a standard multi-layer perceptron (MLP) and the key difference is that it takes the reconstruction error, \emph{i.e.}, the last factor of our representation, as an additional input for each layer. In this way, we avoid the problem of losing the valuable information of reconstruction error through the learning procedure of MLP. The whole system, including the autoencoder and the anomaly detector, is jointly trained by using both the normal data and the few labeled abnormal data.
More details about this system are presented in Section III. 
By conducting experiments on various datasets, we clearly demonstrate the advantage of our anomaly detection model comparing with other state-of-art methods.


The main contributions of this paper are summarized as follows.

\begin{itemize}
\item We introduce a novel strategy to encode the input data, which builds a more effective feature representation that could be used for anomaly detection. 
\end{itemize}

\begin{itemize}
\item Based on our encoding strategy, we propose an anomaly detection system, which seamlessly incorporates the information from the proposed encoding. 
\end{itemize}

\begin{itemize}
\item We conducted an extensive experimental analysis of the proposed method on various datasets and demonstrated the effectiveness of our approach. Ablation studies are also performed to investigate the impact of various components in our method. 
\end{itemize}

The rest part of the paper is organized as follows. Section II introduces the related works. Section III presents the proposed model, including the strategy to encode the input data on the manifold and the proposed anomaly detection method. Extensive experiments and results are provided in Section IV. We conclude the paper in Section V. 


 

\section{Related Works}

\subsection{Deep Supervised Anomaly Detection}

Deep supervised anomaly detection algorithms have been well studied over the last few years. These algorithms outperform the traditional strategies and have made remarkable progress in various application scenarios such as cyber intrusion detection, wireless anomaly detection and etc. In \cite{jiang2018deep}, Jiang et al. proposed a long short-term memory method to detect network attacks. They constructed a multi-channel intelligent attack detection model, which employs multi-channel processing to generate classifiers based on the extracted features of different types and decides the detection results by a voting algorithm to synthesize the outputs of the classifiers. The adoption of deep neural networks demonstrates a significant improvement on the attack detection performance. In \cite{yang2018active}, Yang et al. introduced the human-in-the-loop active learning approach to deal with the intrusion detection problem in wireless Internet of Things. Human experience is considered in the machine learning loop via active learning, which could train the detection model with the most crucial data in an incremental manner. This approach could improve the performance of existing models based on classical machine learning, while reducing the effort of labeling training data. Liu et al.\cite{liu2021deeppayload} designed a specially designed neural network that includes long short-term memory, convolutional neural networks and multi-head self attention mechanism to detect packet payload anomalies. It could learn the potential dependency relationships among the packet payload in both perspectives, \emph{i.e.}, a global perspective as well as a perspective from local features to aggregative representation.

Deep supervised based anomaly detection methods can usually make satisfactory performance when a sufficient number of abnormal data is available and annotated samples are relatively balanced for both data categories. However, supervised approaches often have limited applicability due to the difficulty in collecting anomaly data which is usually much more rare than normal data.

\subsection{Deep Unsupervised Anomaly Detection}

In recent years, deep unsupervised anomaly detection methods have been well studied due to their abilities to capture complex inner relations through feature learning from unlabeled data. In \cite{luo2019video}, Luo et al. proposed an unsupervised anomaly detection method that learns both spatial and temporal features through sparse coding inspired deep neural networks to detect the abnormal events in video data. Their proposed model effectively captures the time coherence for normal events while a real-time detection could be achieved. In \cite{peng2016deep}, Peng et al. proposed a deep learning based subspace clustering method named PARTY to provide the insight for the structure of unlabeled samples, which is vital for unsupervised anomaly detection. PARTY firstly transforms the input data into a nonlinear subspace and leverages both the local and global subspace structure to cluster the data. In \cite{xu2018unsupervised}, Pei et al. proposed a variational autoencoder (VAE) based method named \emph{Donut} to detect anomalies in KPI data (a kind of time series data). Benefit from the effective representation of the deep generative model, the performance of \emph{Donut} is significantly better than those of existing supervised and VAE-based methods. Inspired by robust principal component analysis, Zhou et al.\cite{zhou2017anomaly} proposed a novel deep generative model that could not only effectively discover non-linear features from the original data but also be robust with respect to the outliers and noise in the training data. Based on this model, they also derived a new family of unsupervised anomaly detection methods, which has a superior performance over the baselines. In \cite{zenati2018adversarially}, Zenati et al. explored the direction that employs generative adversarial networks (GANs) to deal with the unsupervised anomaly detection task. They make use of bi-directional GANs to obtain adversarially learned features and determine anomalies based on reconstruction errors. In \cite{zong2018deep}, Zong et al. proposed an unsupervised anomaly detection method called DAGMM, which employs the deep autoencoder and the gaussian mixture model to deal with high-dimensional data. DAGMM leverages a deep autoencoder to extract effective representations in both the original feature space and the low-dimensional latent feature space for each sample, and uses the gaussian mixture model to estimate data density based on the extract feature representations. The two components of DAGMM are jointly optimized in an end-to-end fashion, which leads a significant improvement of detection performance over existing techniques. In\cite{pang2018learning}, Pang et al. proposed an unsupervised anomaly detection framework to learn efficient low-dimensional representations from ultrahigh-dimensional data.  Through unifying the representation learning process and the anomaly detection task, optimal and stable representations for the target detector could be extracted. In \cite{nicolau2018learning}, Cao et al. leveraged effectively latent representations to further improve the detection performance for network anomalies. They proposed two models based on different types of autoencoders. The normal data is forced into a tight area in the latent feature space of their models, which could directly help unsupervised anomaly detection methods to deal with the challenges of high dimensionality and sparsity in network data.

Different from the supervised approaches, unsupervised algorithms learn feature representations and prior knowledge from unlabeled samples based on the hypothesis of data distribution. In many applications, those hypotheses hold and can lead to reasonable anomaly detection performance. However, the unsupervised approaches may face performance degradation problems when the boundary of different data categories is vague or complicated. One of the most important reasons could be that the data distribution modeled by unsupervised approaches may suffer from wrong label assignment. This may only be corrected with labeled anomalous samples.

\subsection{Anomaly Detection With Limited Labels}

To take advantage of the limited but valuable anomalous samples, a few works have made exploration to detect anomalies with extremely unbalanced data in the more recent years. In \cite{siddiqui2018feedback}, Siddiqui et al. proposed a feedback-guided anomaly detection framework. The framework could optimize the unsupervised anomaly detection model by incorporating with the analyst’s prior knowledge to generate higher anomaly scores for instances with higher anomalous probabilities. In \cite{hu2019utilizing}, Hu et al. developed an electricity fraud detection model based on user's consumption patterns. The model effectively extracts consumption patterns from load profiles based on deep neural networks and is trained in a semi-supervised manner through alternate multitask learning. The deep network structure and the training strategy enable the model to have the ability to deal with the challenge of high dimensionality and insufficient labeled anomalies in electricity fraud detection. In \cite{ruff2019deep}, Ruff et al. proposed a deep weakly-supervised anomaly detection method named Deep SAD, which is the extension of a recent released unsupervised anomaly detection, \emph{i.e.}, Deep SVDD \cite{ruff2018deep}, to make use of labeled anomalies. Inspired by the information-theoretic analysis, Deep SAD derives a novel loss to force the normal data to be closed to a fixed center and map the anomalies further away. In\cite{pang2019deepweakly}, Pang et al. proposed an anomaly detection formulation for the scenario with limited labeled abnormal data and abundant unlabeled data. They convert the anomaly detection task into an ordinal regression problem of pairwise relation, which fully utilizes the limited number of labeled anomaly instances to form instance pairs for the follow-up anomaly detection.

The most related work to our paper is \cite{pang2019deep}, which proposed a novel anomaly detection framework called DevNet to detect anomalous instances by empoying only a limited number of labeled anomalies. DevNet achieves an end-to-end learning for anomaly scores that could indicate the anomalous probabilities. It employs a deviation neural network and enforces significant distribution deviations of the anomaly scores between the anomalies and normal data. Different from DevNet, our framework focuses on creating a better feature representation for anomaly detection. This representation, plus the network modules to leverage the representation can be seamlessly incorporated into our framework. In our implementation, we also generate the anomaly score as in \cite{pang2019deep} but with a different loss function.

\section{Proposed Model}
\subsection{System Overview}
\begin{figure*}[!ht]
\centering
\includegraphics[width=0.79\linewidth]{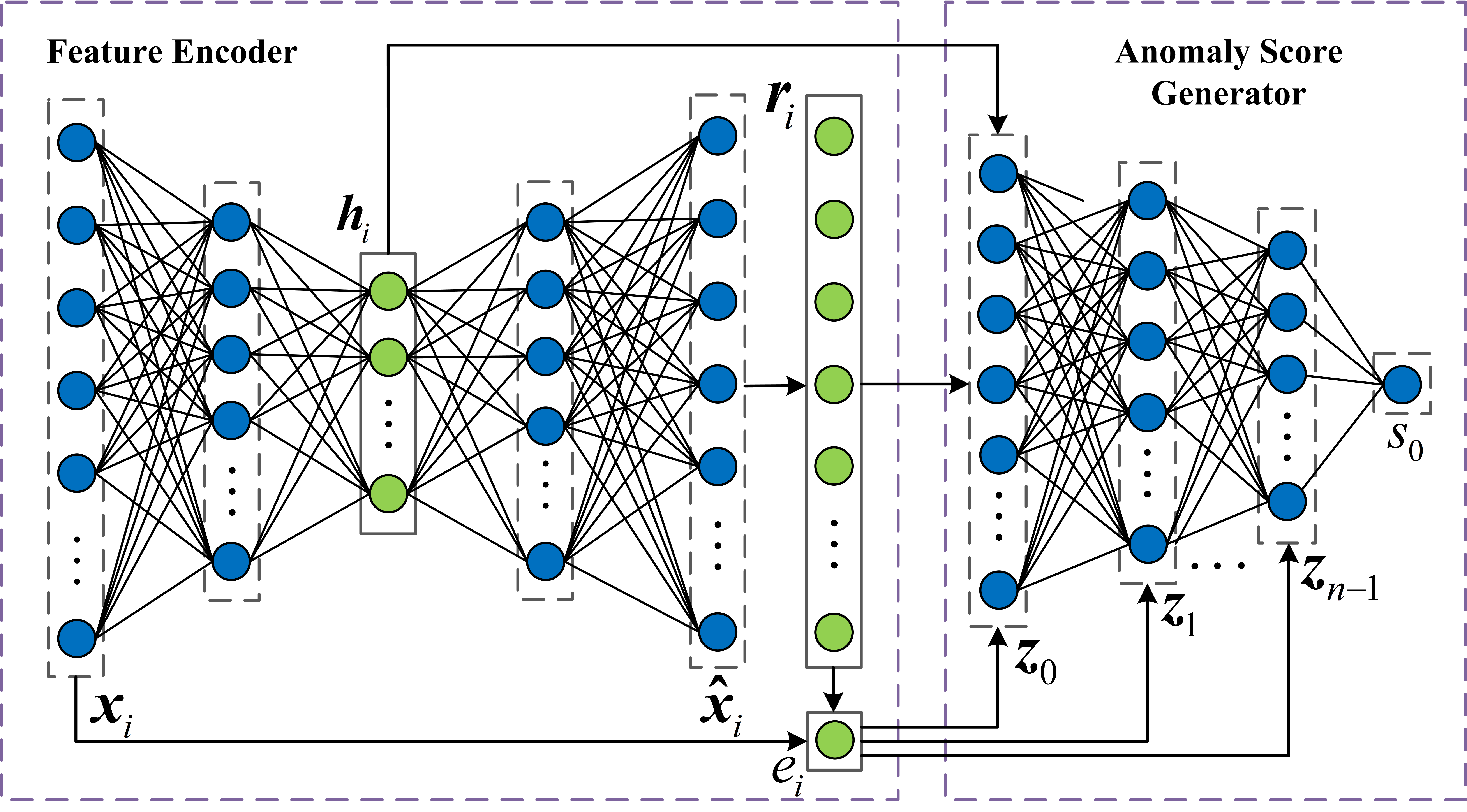}
\centering
\caption{An illustration of the proposed model. The model consists of two modules, \emph{i.e.,} a feature encoder and an anomaly score generator. The feature encoder transforms the input data into a new representation formed by the three factors, \emph{i.e.,} $\bm{h_i}$, $\bm{r_i}$ and $e_i$, which are colored in pale green in the figure. The anomaly score generator calculates a score from the encoded feature representation to indicate the possibility that the input sample is an anomaly.}
\label{threefactor-new}
\end{figure*}

We propose an anomaly detection model with a novel feature encoding strategy, which is designed to cope with the limitation of data insufficiency for annotated anomaly samples. As shown in Figure \ref{threefactor-new}, there are two modules, \emph{i.e.}, a feature encoder and an anomaly score generator, which cooperate with each other in the model. The feature encoder leverages an autoencoder to transform the input data into a more meaningful representation, that is, three factors that characterize how anomaly deviates from the normal pattern.
The anomaly score generator calculates a score for each input sample from the encoded feature representation. 
The feature encoder and the anomaly score generator are jointly optimized through a two-stage training procedure, which will be introduced in Section III-E.
Details of the proposed model are presented as in the following subsections.
\subsection{Feature Encoder}

The feature encoder aims to transform the input data into a new representation that could benefit the anomaly detection task. An autoencoder is leveraged to generate the representation. In particular, three factors, \emph{i.e.}, hidden representation, reconstruction residual vector, and reconstruction error, from the autoencoder are extracted to form the representation. Those three factors characterize how an anomaly sample deviates from the normal pattern. It is expected that anomaly detectors learned on top of those three factors can have better generalization performance on unseen data.

The autoencoder consists of an encoder and a decoder. The encoder first encodes the input data from the original space \( \mathbb{R}^{m}\) to the latent feature space \(\mathbb{L}\subset \mathbb{R}^{d} (d<m)\) and then the decoder converts the latent representation from \(\mathbb{L}\) back to the original space. Let \(\bm{f_e}(\cdot; \bm{W_{e}} ): \mathbb{R}^{m}\mapsto \mathbb{L}\) denote the encoder with
parameters \(\bm{W_{e}}\)
 and \(\bm{f_d} (\cdot; \bm{W_d} ): \mathbb{L}\mapsto \mathbb{R}^m\) be the decoder with parameters \(\bm{W_d}\). 
Given an input sample \(\bm{x}_i \in \mathbb{R}^m\), its latent representation through an autoencoder can be represented as
\begin{equation}
\bm{h}=\bm{f_e} (\bm{x}_i; \bm{W_e} ),
\end{equation}
where \(\bm{h}\in \mathbb{L}\) is a hidden representation based on the autoencoder, which is also the first factor of our feature representation.

Let the reconstruction vector of the autoencoder for \(\bm{x}_i\) be \(\bm{\hat{x}}_i \in \mathbb{R}^m\). \(\bm{\hat{x}}_i \) can be obtained through the decoder:
\begin{equation}
\bm{\hat{x}}_i=\bm{f_d} (\bm{h}; \bm{W_d} ). 
\end{equation}
Based on $\bm{\hat{x}}_i$, two additional factors can be extracted. The first is the reconstruction error, which is defined as 
\begin{align}
    e = \| \bm{\hat{x}}_i - \bm{x}_i \|_2,
\end{align}which is also commonly used as an indicator for anomaly samples.

Then, the second factor of the feature encoding, \emph{i.e.}, the reconstruction residual vector 
can be represented as
\begin{align}
    \bm{r} = \frac{\bm{\hat{x}}_i - \bm{x}_i}{\| \bm{\hat{x}}_i - \bm{x}_i \|_2}.
\end{align}Please note that we normalize the residual vector by the reconstruction error. Thus $\bm{r}$ essentially indicates the direction of original residual vector.

To sum up, the three factors, $\bm{h}$, $e$ and $\bm{r}$ are used to characterize the input sample. We also call them the feature representations of an input sample. Note that the three factors uniquely characterize an input sample. Benefit from this encoding, a typical anomaly could significantly deviate from the normal samples in one of the following cases: large coordinate errors, abnormal reconstruction errors, abnormal residual directions, or any possible combinations of the former three. 


The above three factors also have a geometric implication. In high-dimensional space, data usually resides in a manifold with low intrinsic dimensionality. The encoder builds a nonlinear transform to convert an original high-dimensional data into a more compact low-dimensional feature, which essentially provides the intrinsic coordinate of that data on the manifold. The decoder will transform all points in $\mathbb{L}$ into a manifold $\mathcal{M} \subset \mathbb{R}^m$. If reconstruction error is minimized as an objective to train the autoencoder, $\mathcal{M}$ should be close to the original data manifold. For a new test point $x_t$, the trained autoencoder will map it to a hidden vector $\bm{h}_t \in \mathbb{L}$ and obtain a reconstructed vector $\hat{x}_t \in \mathbb{R}^m$. This process can be seen as {projecting} $x_t$ to $M$. The projection, \emph{i.e.}, the closest point on $\mathcal{M}$ is actually $\hat{x}_t$
\footnote{Strictly speaking, this can only be seen as an approximated closest point.}. This is because if the autoencoder is well trained, it should approximately minimize $\|x_t - \hat{x}_t\|$, which is equivalent to finding the closest point to $x_t$ on $\mathcal{M}$. Therefore, $\bm{h}_t$ can be seen as the intrinsic coordinate of the projection, the corresponding reconstruction error and the residual vector can be seen as the distance and the direction between the projection and the original data point.

\subsection{Anomaly Score Generator}

The anomaly score generator generates an anomaly score to indicate the likelihood of anomaly for an input sample based on the feature representation from the autoencoder. At the training stage, the anomaly score generator is trained to ensure that the score of normal samples follows a prior distribution while the anomaly sample deviates from that. The training process and objective functions will be elaborated in Section III-D. This subsection focuses on how to map the three-factor-representation described above into the anomaly score. 

The most straightforward way is to use a multi-layer perceptron (MLP) to map the concatenation of three factors into a scalar. However, the second factor, \emph{i.e.}, reconstruction error, only has one dimension and is likely to be overpowered by the high-dimensional vectors from the other two factors in the prorogation through multiple layers. To overcome this issue, we propose a solution to add the reconstruction error as an extra dimension to all the layers of MLP. In this way, we could emphasize the importance of the reconstruction error for anomaly detection. The architecture of the anomaly score generator is shown in Figure \ref{threefactor-new}.

Formally, for each layer of the MLP, we have
\begin{align}
    \bm{z}_k = f(\bm{W}_m^k \bm{z}_{k-1} + b_k + w_e^k e),
\end{align}where $\bm{z}_k$ and $\bm{z}_{k-1}$ are the activations at the $k$-th layer and $k-1$-th layer, respectively. $\bm{W}_m^k$ and $b_k$ are the weight and bias for the $k$-th layer. $e$ is used as an extra dimension and $w_e^k$ is its corresponding weight. For the first layer of MLP $\bm{z}_{k-1} = [\bm{r},\bm{h}]$. For the last layer $\bm{z}_k$ becomes the anomaly score $\bm{s}_{0}$.

\subsection{Objective Function}

The proposed model involves two parts, the feature encoder, and the anomaly score generator, which will be jointly optimized through the training process. The objective of training consists of two sub-goals: (1) minimize the reconstruction error of the autoencoder for normal data; (2) ensure that the anomaly score is discriminative for distinguishing anomalies from the normal ones. 


Before elaborating our objective function, let's define the notations that will be used in the following part. The designed network consists of two parts, a feature encoding network \(\psi(\cdot; \bm{\Theta}_e)\) and an anomaly score generator  \(\varphi(\cdot; \bm{\Theta}_g)\), where $\bm{\Theta}_e$ and $\bm{\Theta}_g$ are the model parameters for the feature encoder and anomaly score generator, respectively. Note that the feature encoder is essentially an autoencoder. As described in Section III-B, three factors $\bm{h}_i,\bm{r}_i$ and $e_i$ are extracted from the autoencoder as the representation for the input data $\bm{x}_i$. Those three factors will be fed into \(\varphi(\cdot; \bm{\Theta}_g)\) to produce an anomaly score. The entire system is denoted as $\phi\left(\bm{x}_{i} ; \bm{\Theta}_e, \bm{\Theta}_g\right)$, or $\phi\left(\bm{x}_{i}\right)$ in short.

The first part of the objective function is to minimize the reconstruction error, which is equivalent to minimizing the expectation of $e$. In addition to that, we further encourage that the reconstruction error becomes discriminative for anomaly samples and usual samples. In other words, we expect that the reconstruction error for the anomaly samples should be larger than a threshold. This gives us the following max-marge style loss term:
\begin{equation}
\label{deviationfore}
\begin{split}
\mathcal{L}_e= \sum_i (1-y_i)e_i
+y_i \max \left(0, a_{0}-e_i\right),
\end{split}
\end{equation}where $y_i \in \{0,1\}$ indicates if the current sample is an anomaly sample ($y_i=1$). $a_0$ is a predefined constant (set to 5 in our implementation).

Inspired by \cite{pang2019deep}, we expect that the anomaly score of the normal sample fits a prior distribution while the score for the anomaly deviates from that. We achieve this by using a more simplified loss term than that in \cite{pang2019deep}. Specifically, we expect the anomaly scores of normal samples to be close to 0 while the anomaly scores for outliers deviating from 0 by a predefined margin, which gives the following loss function:

\begin{equation}
\label{devnetloss}
\mathcal{L}_d = \sum_i (1-y_i)\left|\phi\left(\bm{x}_{i}\right)\right|
+y_i \max \left(0, a_{0}-\phi\left(\bm{x}_{i}\right)\right).
\end{equation}

Based on Equation (\ref{deviationfore}) and Equation (\ref{devnetloss}), our joint loss can be formally represented as

\begin{equation}
\label{proposedloss}
\mathcal{L}(\bm{\Theta}_e, \bm{\Theta}_g) =   \mathcal{L}_d(\bm{\Theta}_e, \bm{\Theta}_g) + \lambda\mathcal{L}_e(\bm{\Theta}_e),
\end{equation}where \(\lambda\) is a hyper-parameter to balance the contributions of two parts to the entire loss. Note that $\mathcal{L}_e$ only involves parameters in the feature encoding network. 



Note that the actual annotation of labeled normal samples is not available in the weakly-supervised anomaly detection task. However, most of the unlabeled training samples are normal samples since anomalies are rare in the real world data. To obtain a sufficient number of annotated normal samples for the proposed end-to-end jointly optimization, we employ the same strategy introduced in \cite{pang2019deep}, \emph{i.e.}, simply treating all unlabeled training samples as normal samples. Through our experiments, we show that this strategy achieves satisfactory performance.

\subsection{Training procedure}

We use stochastic gradient descent to train the network. To balance the abnormal samples and normal samples, we construct a mini-batch by sampling the same amount of abnormal samples and normal samples. In other words, the abnormal samples are over-sampled.

Empirically, we find beneficial effects to utilize a two-stage training strategy to learn the model. The training procedure is detailedly presented in Algorithm 1. We first employ a pre-training stage which trains the feature encoding network by only using the reconstruction loss:
\begin{equation}
\label{reconsloss}
\begin{split}
\mathcal{L}_{ae}(\bm{\Theta}_{e})= \sum_i e_i.
\end{split}
\end{equation}
In this stage, the feature encoding network $\psi(\cdot;\bm{\Theta}_e)$
is pre-trained with all unlabelled training samples, which aims to obtain a basic encoding strategy for the following end-to-end training of the entire network. Note that a root mean square error loss is adopted in this training process. The corresponding procedures in Algorithm 1 are Step 2-6.

Then we train the whole network $\phi(\cdot;\bm{\Theta}_e,\bm{\Theta}_g)$ with the parameters in the feature encoding network being initialized by the parameters learned from the pre-training stage. In the second training stage, all unlabelled data and a few annotated anomalies are used to perform the end-to-end jointly optimization for the entire network based on the pre-training of the feature encoding network. The corresponding procedures in Algorithm 1 are Step 7-12.


\begin{algorithm}[!h]
  \caption{Network Training Procedure}
  \label{alg1}
  \renewcommand{\algorithmicrequire}{\textbf{Stage 1: Pre-training}}
  \begin{algorithmic}[1]
  \REQUIRE
  \renewcommand{\algorithmicrequire}{\textbf{Input:}}\renewcommand{\algorithmicensure}{\textbf{Output:}}
  \REQUIRE $ \bm{X}^{\text {unlabeled}} \in \mathbb{R}^{m}$ 
  \ENSURE $\psi(\cdot;\bm{\Theta}_e):\mathbb{R}^{m}\mapsto\mathbb{R}^{m} $ 
  \STATE Randomly initialize $\bm{\Theta}_e$
  \REPEAT
  \STATE Randomly sample one batch of unlabeled data.
  \STATE Compute the loss $\mathcal{L}_{u}$ and update weight matrices of $\psi(\cdot;\bm{\Theta}_e)$ 
  \UNTIL converge
  \RETURN $\psi(\cdot;\bm{\Theta}_e)$
  \renewcommand{\algorithmicrequire}{\textbf{Stage 2: End-to-end Optimization}}
  \REQUIRE
  \renewcommand{\algorithmicrequire}{\textbf{Input:}}\renewcommand{\algorithmicensure}{\textbf{Output:}}
  \REQUIRE $ \bm{X} \in \mathbb{R}^{m}$ 
  \ENSURE $\phi(\cdot;\bm{\Theta}_e;\bm{\Theta}_g):\mathbb{R}^{m}\mapsto\mathbb{R}$ 
  \STATE Randomly initialize $\bm{\Theta}_g$ and load $\bm{\Theta}_e$ from the pre-trained network $\psi(\cdot;\bm{\Theta}_e)$
  \REPEAT
  \STATE Randomly sample half of samples from unlabeled data and half of samples from labeled anomalies to form a batch
 \STATE Compute the joint loss $\mathcal{L}$ and update the network parameters $\left\{\bm{\Theta}_e;\bm{\Theta}_g\right\}$ 
 \UNTIL converge
 \RETURN $\phi(\cdot;\bm{\Theta}_e;\bm{\Theta}_g)$
\end{algorithmic}
\end{algorithm} 

\section{Experiments}
In this section, we evaluate the performance of the proposed method with extensive experiments. Firstly, we introduce the datasets and performance metrics that are used in the experiments. Then, we present the main results of the proposed algorithm. Finally, we conduct the ablative analysis to investigate the impact of various components in our method.

\subsection{Datasets Description}
In the experiments, we evaluate the proposed method on eight datasets, which were collected from different application scenarios such as network security, finance fraud detection, medical diagnosis and etc. Anomalies from these datasets are defined according to either the domain-specific criteria or the dramatic difference from the majority of the data. Specifically, \emph{NSL-KDD}\cite{tavallaee2009detailed} is a dataset from the network security field, which was extracted from the KDDCUP99 dataset. The network attacks are regarded as anomalies in this dataset.  \emph{Spambase} and \emph{Arrhythmia} are two datasets from the UCI machine learning repository\cite{lichman2013uci}. The spam samples are treated as anomalies in the \emph{Spambase} dataset, and the classes with the fewest number of samples, \emph{i.e.}, 3, 4, 5, 7, 8, 9, 14, and 15, are combined to form the anomalies in the \emph{Arrhythmia} dataset. \emph{Fraud}\cite{dal2017credit} is a dataset from Kaggle, in which the task is to detect credit card fraud. The fraudulent transactions are treated as anomalies. The other four datasets, \emph{i.e.}, \emph{Cardio}, \emph{Shuttle}, \emph{Satellite} and \emph{Mammography}, are from the ODDS library\cite{rayana2016odds}. The \emph{Cardio} is a cardio disease detection dataset, in which the anomalies are the patients diagnosed with the disease. The \emph{Shuttle}, \emph{Satellite} and \emph{Mammography} are three anomaly detection datasets which are originally for multi-class classification problems in different domains, \emph{i.e.}, space shuttle, satellite image and medical image. The combination of the smallest classes in the corresponding classification task forms the “anomaly” class in those three datasets. 

Before employing these datasets to evaluate the anomaly detectors, a standard preprocessing procedure has been performed on each dataset. Firstly, the missing values are filled with the mean value of the corresponding attribute. Then, one-hot encoding is performed on all categorical attributes. Finally, each attribute is normalized into [0, 1]. Details of the datasets are shown in Table \ref{tabledataset}. \textbf{\emph{D}} is the dimension of the pre-processed data. \textbf{\emph{N}} is the number of instances in the dataset. $\bm{f_1}$ denotes the percentage of labeled anomalies in the training datasets.

\begin{table}[!h]
\scriptsize
\caption{\centering Datasets for Evaluating Models.}
\centering
\begin{spacing}{1.15}
\setlength{\tabcolsep}{5mm}{
\begin{tabular}{||c|ccc||}
\hline
\textbf{Dataset} &\textbf{\emph{D}} &\textbf{\emph{N}} & $\bm{f_1}$   \\
\hline
NSL-KDD & 122 & 148,517 & 0.05\% \\
Spambase & 57 & 4,601 &  1.30\%\\
Arrhythmia  & 279 & 452 & 4.56\%\\
Cardio & 21 & 1,831 & 2.17\% \\
Shuttle & 9 & 49,097 & 0.08\% \\
Satellite & 36 & 6,435 & 0.83\% \\
Mammography & 6 & 11,183 &  0.34\% \\
Fraud & 29 & 284,807 & 0.01\% \\
\hline
\end{tabular}}
\end{spacing}
\label{tabledataset}
\end{table}

\subsection{Performance Metrics}
We evaluate the anomaly detectors in the experiments with two performance metrics, \emph{i.e.}, the Area Under Receiver Operating Characteristic Curve (AUC-ROC) and the Area Under Precision-Recall Curve (AUC-PR). AUC-ROC is the area under ROC curve which describes the relationship between true positive rates and false-positive rates, and the AUC-PR is the area under PR curve which represents the relationship between precision and recall. While AUC-ROC evaluates the 

\begin{table*}[!h]
\scriptsize 
\caption{\centering Experimental results (Mean $\pm$ standard deviation) for the Proposed Method and Competitive Methods. The average performance for each method is summarized in the last row. The best result for each row is in bold.}
\centering
\begin{spacing}{1.15}
\setlength{\tabcolsep}{1mm}{
\begin{tabular}{||c|ccccc|ccccc||}
\hline
\textbf{\multirow{2}{*}{ Dataset}}& \multicolumn{5}{c|}{\textbf{ AUC-ROC Performance}}  &\multicolumn{5}{c||}{\textbf{AUC-PR Performance}} \\
\cline{2-6}  \cline{7-11}& \textbf{Proposed}& \textbf{DevNet} &\textbf{Deep SAD} &\textbf{DAGMM} &\textbf{FCN}  & \textbf{Proposed}& \textbf{DevNet} &\textbf{Deep SAD} &\textbf{DAGMM} &\textbf{FCN} \\
\hline
NSL-KDD & 0.955$\pm$0.007 & 0.952$\pm$0.018 & \textbf{0.969}$\pm$0.002 & 0.745$\pm$0.068 &  0.885$\pm$0.132 & \textbf{0.970}$\pm$0.003 & 0.946$\pm$0.012 & 0.965$\pm$0.002 & 0.694$\pm$0.064  & 0.879$\pm$0.149 \\

Spambase & \textbf{0.921}$\pm$0.016 & 0.890$\pm$0.001 & 0.898$\pm$0.004 & 0.530$\pm$0.047  & 0.903$\pm$0.097   & \textbf{0.910}$\pm$0.036 & 0.828$\pm$0.002 & 0.890$\pm$0.005&  0.423$\pm$0.037 & 0.875$\pm$0.123 \\

Arrhythmia & 0.797$\pm$0.048 & 0.772$\pm$0.041 & \textbf{0.845}$\pm$0.019 & 0.728$\pm$0.008  & 0.658$\pm$0.141 & \textbf{0.681}$\pm$0.038 & 0.659$\pm$0.031 & 0.568$\pm$0.069&  0.399$\pm$0.102 & 0.488$\pm$0.184 \\

Cardio & \textbf{0.999}$\pm$0.001 & 0.985$\pm$0.001 &  0.994$\pm$0.002 & 0.885$\pm$0.033 & 0.857$\pm$0.181 & \textbf{0.990}$\pm$0.003 & 0.934$\pm$0.003 & 0.968$\pm$0.008 &  0.623$\pm$0.086 & 0.758$\pm$0.254  \\

Shuttle & 0.983$\pm$0.004 & 0.976$\pm$0.001 & \textbf{0.996}$\pm$0.001 & 0.990$\pm$0.002 &  0.876$\pm$0.001 & \textbf{0.975}$\pm$0.001 & 0.968$\pm$0.001 & 0.952$\pm$0.003 & 0.959$\pm$0.013 & 0.869$\pm$0.001 \\

Satellite & \textbf{0.918}$\pm$0.007 & 0.882$\pm$0.005 &  0.863$\pm$0.010 & 0.739$\pm$0.031 & 0.862$\pm$0.018 & \textbf{0.895}$\pm$0.006 & 0.856$\pm$0.005 & 0.833$\pm$0.007 &  0.684$\pm$0.021 & 0.819$\pm$0.021 \\

Mammography & \textbf{0.972}$\pm$0.002 & 0.969$\pm$0.001 & 0.930$\pm$0.005 & 0.759$\pm$0.079 & 0.793$\pm$0.317 & \textbf{0.645}$\pm$0.021 & 0.642$\pm$0.008 & 0.591$\pm$0.046 &  0.125$\pm$0.070 & 0.459$\pm$0.273 \\

Fraud & \textbf{0.982}$\pm$0.003 & 0.977$\pm$0.001 & 0.946$\pm$0.009  & 0.916$\pm$0.026 & 0.841$\pm$0.251 & \textbf{0.692}$\pm$0.006 & 0.688$\pm$0.004 & 0.563$\pm$0.046 &  0.442$\pm$0.160 & 0.492$\pm$0.265 \\
\hline
Average & \textbf{0.941}$\pm$0.011 & 0.925$\pm$0.009 & 0.930$\pm$0.007 & 0.787$\pm$0.037 & 0.834$\pm$0.142 & \textbf{0.845}$\pm$0.014 & 0.815$\pm$0.008 & 0.791$\pm$0.023 &  0.544$\pm$0.069 & 0.704$\pm$0.102 \\

\hline
\end{tabular}}
\end{spacing}
\label{tablemainresults}
\end{table*}

\noindent prediction performance for both the normal class and the abnormal class, AUC-PR is a performance metric that pays more attention to the anomalies. Note that both AUC-ROC and AUC-PR lie in [0, 1], and a higher value indicates a better performance. We follow the literature\cite{campos2016evaluation} to compute AUC-ROC and calculate AUC-PR according to the literature \cite{manning2008introduction}. All AUC-ROC and AUC-PR results with $\pm$ standard deviations are calculated over ten independent runs.

\subsection{Experiment Results with Competitive Methods}

\emph{\textbf{Experiment Design.}}In this subsection, we conduct experiments to compare the proposed method against other competitive methods. Specifically, four methods, including DevNet\cite{pang2019deep}, Deep SAD\cite{ruff2019deep}, DAGMM\cite{zong2018deep} and a baseline approach which directly uses the binary classification loss to train the same network as DevNet, denoted as FCN. Among those methods, DevNet is the weakly-supervised anomaly detection approach mentioned in Section II-C. It is most related to our method. Deep SAD is also a weakly-supervised anomaly detection approach which extends an unsupervised anomaly detector, \emph{i.e.}, Deep SVDD\cite{ruff2018deep}, to the weakly-supervised setting. DAGMM is a state-of-art unsupervised anomaly detection method. We include FCN as a supervised anomaly detection baseline. It directly treats the weakly-supervised anomaly detection as a supervised learning problem, that is, using labeled anomaly data as positive class samples and unlabeled data as negative class samples. 

We evaluate the proposed method as well as the competitive methods on eight real-world datasets that have been introduced in Section IV-A. We follow the same experimental settings used in the literature \cite{pang2019deep} to construct the labeled set and unlabeled set. The anomalies and normal samples in each dataset are firstly split into two parts, \emph{i.e.}, 80\% samples as the training data and 20\% samples as the testing data.  We randomly select 30 anomalies from the abnormal class in each training dataset such that the labeled anomalies only account for 0.01\%-2.17\% of the training data.  For the Arrhythmia dataset, only 15 labeled anomalies(4.56\% of the training data) are randomly selected due to its small data size. As there are usually a few anomalies among the unlabeled data in realistic scenarios, we also randomly add the 2\% anomaly samples, \emph{i.e.}, 2\% anomaly contamination, to the normal class in each training dataset. The hyper-parameter $a_0$ and $\lambda$ are set to 5 and 1, respectively.

\emph{\textbf{Result Analysis.}} The comparison of the proposed method against the other four competing methods is shown in Table \ref{tablemainresults}. As seen, comparing with the unsupervised method DAGMM and the fully supervised method FCN, the proposed method greatly outperforms them on all the datasets in both AUC-ROC and AUC-PR. Comparing with the other two state-of-art weakly-supervised anomaly detectors, \emph{i.e.}, DevNet and Deep SAD, the proposed method obtains the best results on five and eight datasets in AUC-ROC and AUC-PR respectively. Though the proposed method ranks second on Arrhythmia and Shuttle in AUC-ROC, it achieves much better performance than the competitors on these two datasets in AUC-PR. The average performance of the proposed method surpasses the competing methods by 1.6\% (DevNet), 1.1\% (Deep SAD), 15.4\% (DAGMM), 10.7\% (FCN) in AUC-ROC, respectively, while the respective average improvements on AUC-PR are 3.0\% (DevNet), 5.4\% (Deep SAD), 30.1\% (DAGMM), 14.1\% (FCN), which are more significant than those in AUC-ROC. Overall, the proposed method attains the best average performance among all competing methods. The above results clearly demonstrated the advantage of the proposed encoding method since it is the major difference from the competing methods, \emph{e.g.}, the architecture and loss function of our method are very similar to DevNet if the encoding submodule and the corresponding loss term are removed.



\subsection{Sample Efficiency}
\emph{\textbf{Experiment Design.}} To investigate the sample efficiency of the proposed method, \emph{i.e.}, the impact of using a different number of labeled anomalies in training data, we evaluate the performance of the models trained by varying the amount of labeled abnormal data. In the experiments, the number of available labeled anomalies is set to one of the following values: 30, 60, 90, 120. Experiments are not conducted on Arrhythmia due to the limited number of abnormal samples available. The other experimental settings remain the same as those described in Section IV-C.
\begin{figure*}[!h]
\centering
\includegraphics[width=0.85\linewidth]{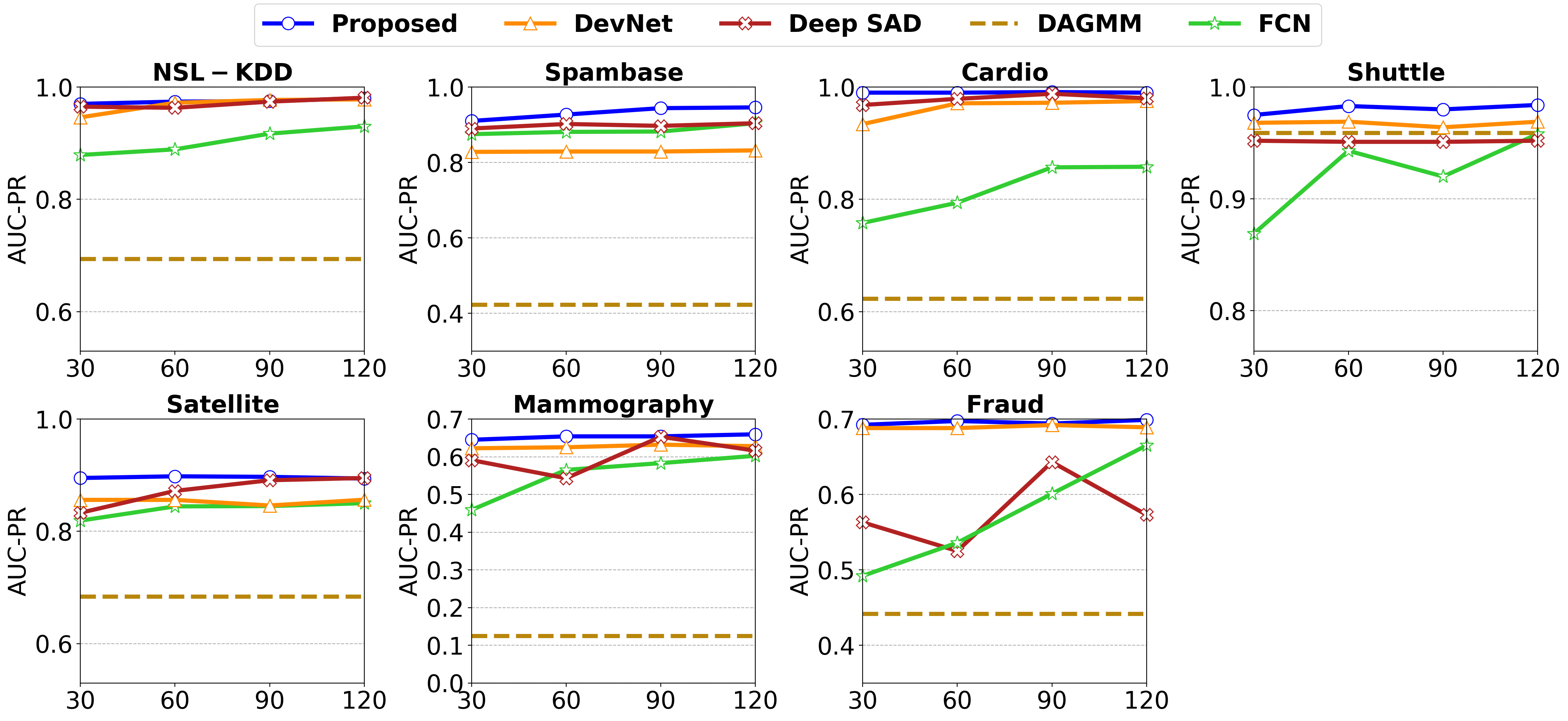}
\centering
\caption{AUC-PR with different number of labeled anomalies for training. The results on Arrhythmia are omitted due to the limited number of abnormal samples available.}
\label{fig-efficiency}
\end{figure*}

\begin{figure*}[!h]
\centering
\includegraphics[width=0.85\linewidth]{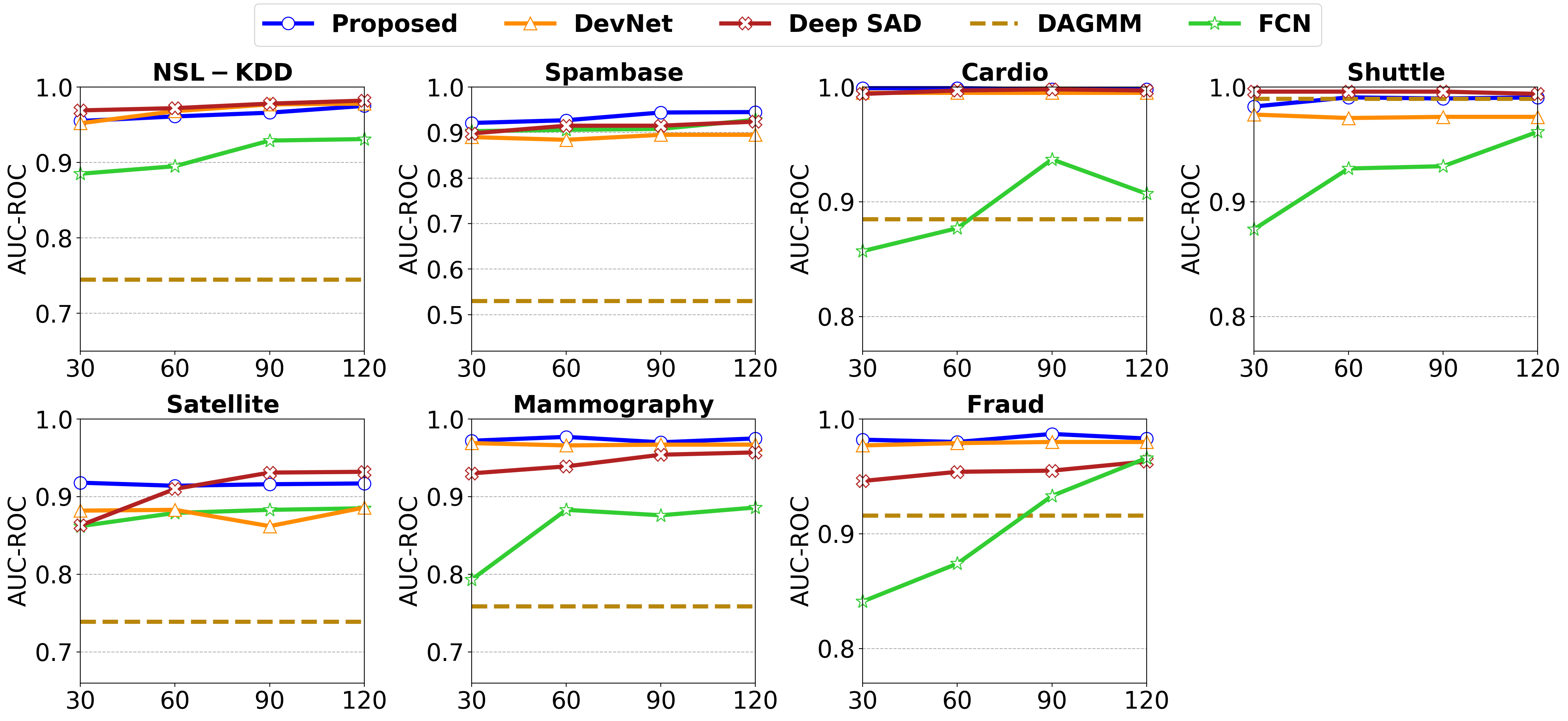}
\centering
\caption{AUC-ROC with different number of labeled anomalies for training. The results on Arrhythmia are omitted due to the limited number of abnormal samples available.}
\label{fig-efficiency-roc}
\end{figure*}

\emph{\textbf{Result Analysis.}}
As shown in Figure \ref{fig-efficiency} and Figure \ref{fig-efficiency-roc}, the proposed method demonstrates significantly better sample efficiency than all the competing methods on AUC-PR. The respective performance of the proposed method on AUC-ROC is comparably well to those of the compared methods. \\

(1) The performance of all the three weakly-supervised anomaly detection methods, \emph{i.e.}, DevNet, Deep SAD, and the proposed method, achieves better performance with the increase of labeled anomalies in general. When there are very few labeled anomaly samples, \emph{i.e.}, 30 or 60, the performance of the proposed method is always the best on AUC-PR for all datasets, while its respective performance on AUC-ROC is comparable to the best performance of the competitors. Also, we can observe that some anomaly detectors can be very sensitive to the quantity change of labeled samples. We could even observe that the performance of some models drops when the number of labeled anomalies increases. In comparison, the curves of our method are more stable and do not exhibit an obvious drop with the increase of labeled data. \\
\indent(2) Compared with the fully supervised method, \emph{i.e.}, FCN, the proposed method can always achieve superior performance (for both AUC-ROC and AUC-PR) by employing the same number of labeled anomalies for training. This observation indicates that the proposed method can consistently achieve improvement by employing the unlabeled data and a few labeled anomalies. \\
\indent(3) Compared with the unsupervised method, \emph{i.e.}, DAGMM, the proposed method achieves significantly better performance even when a very limited number of labeled anomalies are used to provide the prior knowledge. For example, the average performance of the proposed method employing 30 labels surpasses that of DAGMM by more than 30.39\% (AUC-PR) and 16.66\% (AUC-ROC). This also shows that weakly-supervised anomaly detection is a more effective anomaly detection strategy and one can obtain significant improvement by even using a small amount of labeled data.

\begin{table*}[!h]\scriptsize
\caption{\centering Experimental results (Mean $\pm$ standard deviation) to investigate on the effect of the encoding strategy (the first group of experiments to evaluate the effectiveness of the three factors).}
\centering
\begin{spacing}{1.15}
\setlength{\tabcolsep}{1.8mm}{
\begin{tabular}{||c|ccccccc||}
\hline
\textbf{\multirow{2}{*}
{Dataset}}& \multicolumn{7}{c||}{\textbf{AUC-ROC Performance}}  \\
\cline{2-8}  & $\bm{h}$ & $\bm{r}$ & $e$  &$\bm{h}$, $\bm{r}$  & $\bm{h}$, $e$ & $\bm{r}$, $e$  & $\bm{h}$, $\bm{r}$, $e$ \\
\hline

NSL-KDD & 0.926$\pm$0.003 & 0.936$\pm$0.008 & 0.943$\pm$0.005 & 0.943$\pm$0.006 & 0.954$\pm$0.006 & 0.931$\pm$0.012 & \textbf{0.955}$\pm$0.007 \\
Spambase & 0.893$\pm$0.008 & 0.852$\pm$0.016 & 0.898$\pm$0.013 & 0.895$\pm$0.013 & 0.906$\pm$0.012 & 0.869$\pm$0.016 & \textbf{0.921}$\pm$0.016 \\
Arrhythmia & 0.771$\pm$0.035 & 0.752$\pm$0.048 & 0.756$\pm$0.055 & 0.769$\pm$0.055 & 0.764$\pm$0.053 & 0.725$\pm$0.064 & \textbf{0.797}$\pm$0.048  \\
Cardio & 0.996$\pm$0.000 & 0.997$\pm$0.001 & 0.997$\pm$0.000 & 0.998$\pm$0.000 & \textbf{0.999}$\pm$0.000 & 0.998$\pm$0.000 & \textbf{0.999}$\pm$0.001 \\
Shuttle & 0.981$\pm$0.006 & 0.982$\pm$0.004 & 0.981$\pm$0.004 & 0.982$\pm$0.003 & 0.979$\pm$0.010 & 0.982$\pm$0.008 & \textbf{0.983}$\pm$0.004 \\
Satellite & 0.879$\pm$0.004 & 0.885$\pm$0.006 & 0.869$\pm$0.008 & 0.887$\pm$0.005 & 0.874$\pm$0.004 & 0.874$\pm$0.007 & \textbf{0.918}$\pm$0.007 \\
Mammography & 0.968$\pm$0.003 & 0.966$\pm$0.007 & 0.890$\pm$0.011 & 0.970$\pm$0.002 & 0.915$\pm$0.006 & 0.958$\pm$0.017 & \textbf{0.972}$\pm$0.002 \\
Fraud & 0.967$\pm$0.004 & 0.966$\pm$0.010 & 0.949$\pm$0.011  & 0.963$\pm$0.003 & 0.960$\pm$0.007 & 0.960$\pm$0.006 & \textbf{0.982}$\pm$0.003  \\
\hline
Average & 0.916$\pm$0.008 & 0.917$\pm$0.013 & 0.910$\pm$0.013  & 0.926$\pm$0.011 & 0.919$\pm$0.012 & 0.912$\pm$0.016 & \textbf{0.941}$\pm$0.011  \\
\hline
\hline
\textbf{\multirow{2}{*}
{Dataset}}& \multicolumn{7}{c||}{\textbf{AUC-PR Performance}}  \\
\cline{2-8}  & $\bm{h}$ & $\bm{r}$ & $e$  &$\bm{h}$, $\bm{r}$  & $\bm{h}$, $e$ & $\bm{r}$, $e$  & $\bm{h}$, $\bm{r}$, $e$ \\
\hline
NSL-KDD & 0.969$\pm$0.002 & 0.960$\pm$0.003 & 0.964$\pm$0.003 & 0.965$\pm$0.003 & \textbf{0.970}$\pm$0.004 & 0.958$\pm$0.006 & \textbf{0.970}$\pm$0.003 \\
Spambase & 0.901$\pm$0.004 & 0.875$\pm$0.009 & 0.901$\pm$0.007 & 0.902$\pm$0.008 & 0.892$\pm$0.009 & 0.884$\pm$0.011 & \textbf{0.910}$\pm$0.036 \\
Arrhythmia  & 0.663$\pm$0.032 & 0.650$\pm$0.033 & 0.623$\pm$0.043 & 0.650$\pm$0.029 & 0.676$\pm$0.036 & 0.613$\pm$0.056 & \textbf{0.681}$\pm$0.038  \\
Cardio & 0.989$\pm$0.003 & 0.989$\pm$0.004 & 0.989$\pm$0.003 & 0.987$\pm$0.003 & 0.987$\pm$0.002 & 0.989$\pm$0.002 & \textbf{0.990}$\pm$0.003 \\
Shuttle & 0.973$\pm$0.005 & 0.971$\pm$0.004 & 0.966$\pm$0.005 & 0.973$\pm$0.001 & 0.966$\pm$0.006 & 0.971$\pm$0.006 & \textbf{0.975}$\pm$0.001 \\
Satellite & 0.851$\pm$0.003 & 0.853$\pm$0.005 & 0.829$\pm$0.011 & 0.860$\pm$0.004 & 0.842$\pm$0.008 & 0.853$\pm$0.006 & \textbf{0.895}$\pm$0.006 \\
Mammography & 0.616$\pm$0.011 & 0.601$\pm$0.037 & 0.273$\pm$0.007 & 0.615$\pm$0.018 & 0.361$\pm$0.039 & 0.602$\pm$0.086 & \textbf{0.645}$\pm$0.021 \\
Fraud & 0.689$\pm$0.005 & 0.691$\pm$0.007 & 0.671$\pm$0.026  & 0.686$\pm$0.006 & 0.686$\pm$0.007 & 0.687$\pm$0.005 & \textbf{0.692}$\pm$0.006 \\
\hline
Average & 0.831$\pm$0.008 & 0.824$\pm$0.013 & 0.777$\pm$0.013  & 0.830$\pm$0.009 & 0.798$\pm$0.014 & 0.820$\pm$0.009 & \textbf{0.845}$\pm$0.014  \\
\hline
\end{tabular}}
\end{spacing}
\label{tableforq}
\end{table*}

\subsection{Ablative Analysis}
In this subsection, we perform ablative analysis to explore the effects of different components on the performance of our proposed method. Specially, we create variants of our method to investigate the effects of the encoding strategy, the objective function, the hyper-parameters, the pre-training procedure and the strength of the reconstruction error, respectively.

\subsubsection{Effect of Encoding Strategy}

To investigate the impact of the encoding strategy, we conduct two groups of experiments. The first group of experiments aims to evaluate the effectiveness of the three factors, \emph{i.e.}, $\bm{h}$, $\bm{r}$ and $e$, in the feature encoder. There are six variants of models that include different combinations of the three factors (as shown in Table \ref{tableforq}). The resultant encodings are fed into the anomaly score generator. Note that if a model generates the factor $e$, \emph{i.e.}, the reconstruction error, this scalar value is added to each layer of the anomaly score generator as an additional dimension of input (as described in Section III-C). The other experimental settings of the six models remain the same as that of the proposed method described in Section IV-C.

The experiment results show that the three factors are all indispensable in the proposed model. From Table \ref{tableforq}, we can see that the proposed method achieves the best performance among those variants that only use a single factor or two combined factors. The average performance of the model employing all the three factors is 1.5\%-3.1\% and 1.4\%-6.8\% higher than those of the compared models in AUC-ROC and AUC-PR, respectively. 

The second group of experiments is designed to examine whether the performance improvement of our proposed model comes from additional parameters included in the feature encoder. We compare a model that adopts the same network structure as the original one in the feature encoder. However, instead of outputting the three factors in the original feature encoder, the alternative encoder just outputs the reconstructed vector, \emph{i.e.}, $\bm{\hat{x}}_i$ as defined in Equation (2), to the downstream anomaly score generator. Note that the network structure of the anomaly score generator degenerates into an MLP in this case. The other experimental settings of the compared model remain the same as that of the proposed method. 

\begin{table}[!h]\scriptsize
\caption{\centering Experimental Results (Mean $\pm$ standard deviation) to investigate on the effect of the encoding strategy (the second group of experiments to examine whether the performance improvement for the proposed model comes from additional parameters included in the feature encoder).}
\centering
\begin{spacing}{1.15}

\setlength{\tabcolsep}{1.5mm}{
\begin{tabular}{||c|cc|cc||}
\hline
\textbf{\multirow{2}{*}
{Dataset}}& \multicolumn{2}{c|}{\textbf{AUC-ROC Performance}} &\multicolumn{2}{c||}{\textbf{AUC-PR Performance}}  \\
\cline{2-3} \cline{4-5} & \textbf{Proposed} & \textbf{Variant} & \textbf{Proposed} & \textbf{Variant}\\
\hline

NSL-KDD & \textbf{0.955}$\pm$0.007 & 0.943$\pm$0.012 & \textbf{0.970}$\pm$0.003 & 0.965$\pm$0.006  \\
Spambase & \textbf{0.921}$\pm$0.016 & 0.834$\pm$0.028 & \textbf{0.910}$\pm$0.036 & 0.854$\pm$0.018 \\
Arrhythmia  & \textbf{0.797}$\pm$0.048 & 0.784$\pm$0.049 & \textbf{0.681}$\pm$0.038 & 0.638$\pm$0.054  \\
Cardio & \textbf{0.999}$\pm$0.001 & 0.990$\pm$0.011 & \textbf{0.990}$\pm$0.003 & 0.961$\pm$0.028  \\
Shuttle & \textbf{0.983}$\pm$0.004 & 0.975$\pm$0.007 & \textbf{0.975}$\pm$0.001 & 0.961$\pm$0.007 \\
Satellite & \textbf{0.918}$\pm$0.007 & 0.875$\pm$0.013 & \textbf{0.895}$\pm$0.006 & 0.847$\pm$0.013  \\
Mammography & \textbf{0.972}$\pm$0.002 & 0.928$\pm$0.039 & \textbf{0.645}$\pm$0.021 & 0.489$\pm$0.124  \\
Fraud & \textbf{0.982}$\pm$0.003 & 0.954$\pm$0.014 & \textbf{0.692}$\pm$0.006  & 0.678$\pm$0.023  \\
\hline
Average & \textbf{0.941}$\pm$0.011 & 0.910$\pm$0.022 & \textbf{0.845}$\pm$0.014  & 0.799$\pm$0.034  \\
\hline

\end{tabular}
}
\end{spacing}
\label{tableaemlp}
\end{table}

As shown in Table \ref{tableaemlp}, the proposed method achieve substantial performance improvement on all the datasets, \emph{i.e.}, average improvement of 3.2\% and 3.0\% in AUC-ROC and AUC-PR, respectively. This experimental result clearly demonstrates that the performance improvement of the proposed method is not a result of the increase in network parameters.

\subsubsection{Effect of Having Reconstruction Error Regularization Term on Anomaly Score Generator}

We also investigate the impact of using reconstruction error as the regularization term in our objective function. We compare the performance of the proposed method with a variant version which does not enforce the reconstruction error $e$ minimization in Equation (\ref{proposedloss}).  Again, we fix the other experimental settings to make a fair comparison.

\begin{table}[!h]\scriptsize
\caption{\centering Experimental results (Mean $\pm$ standard deviation) to investigate the Effect of Having Reconstruction Error Regularization Term on Anomaly Score Generator. }
\centering
\begin{spacing}{1.15}
\setlength{\tabcolsep}{1.2mm}{
\begin{tabular}{||c|cc|cc||}
\hline
\textbf{\multirow{2}{*}
{Dataset}}& \multicolumn{2}{c|}{\textbf{AUC-ROC Performance}} &\multicolumn{2}{c||}{\textbf{AUC-PR Performance}}  \\
\cline{2-3} \cline{4-5} & \textbf{Proposed Loss} & \textbf{Variant Loss} & \textbf{Proposed Loss} & \textbf{Variant Loss}\\
\hline
NSL-KDD & \textbf{0.955}$\pm$0.007 & 0.954$\pm$0.007 & \textbf{0.970}$\pm$0.003 & 0.969$\pm$0.003  \\
Spambase & \textbf{0.921}$\pm$0.016 & 0.893$\pm$0.013 & \textbf{0.910}$\pm$0.036 & 0.901$\pm$0.007 \\
Arrhythmia  & \textbf{0.797}$\pm$0.048 & 0.771$\pm$0.050 & \textbf{0.681}$\pm$0.038 & 0.663$\pm$0.042  \\
Cardio & \textbf{0.999}$\pm$0.001 & 0.998$\pm$0.001 & \textbf{0.990}$\pm$0.003 & 0.989$\pm$0.004  \\
Shuttle & \textbf{0.983}$\pm$0.004 & 0.981$\pm$0.004 & \textbf{0.975}$\pm$0.001 & 0.972$\pm$0.002 \\
Satellite & \textbf{0.918}$\pm$0.007 & 0.879$\pm$0.006 & \textbf{0.895}$\pm$0.006 & 0.851$\pm$0.006  \\
Mammography & \textbf{0.972}$\pm$0.002 & 0.968$\pm$0.005 & \textbf{0.645}$\pm$0.021 & 0.616$\pm$0.018  \\
Fraud & \textbf{0.982}$\pm$0.003 & 0.967$\pm$0.003 & \textbf{0.692}$\pm$0.006  & 0.689$\pm$0.009  \\
\hline
Average & \textbf{0.941}$\pm$0.011 & 0.926$\pm$0.011 & \textbf{0.845}$\pm$0.014  & 0.831$\pm$0.011  \\
\hline
\end{tabular}}
\end{spacing}
\label{tableloss}
\end{table}

As shown in Table \ref{tableloss}, comparing with the model using the original objective function, the model employing this variant objective function has led to an average performance drop of 2.3\% and 2.3\% in AUC-ROC and AUC-PR, respectively. This experimental result demonstrates that the reconstruction error regularization term on anomaly score generator plays a crucial role.


\begin{table*}[!hb]\scriptsize
\caption{\centering Experimental results (Mean $\pm$ standard deviation) to investigate on the effects of the hyper-parameter $a_0$.}
\centering
\begin{spacing}{1.15}
\setlength{\tabcolsep}{1.8mm}{
\begin{tabular}{||c|ccccccccc||}
\hline
\textbf{\multirow{2}{*}
{Dataset}}& \multicolumn{9}{c||}{\textbf{AUC-ROC Performance}}  \\
\cline{2-10} & $a_0=0.1$ & $a_0=1$ & $a_0=3$ & $a_0=4$ & $a_0=5$ & $a_0=6$ & $a_0=7$ & $a_0=10$ & $a_0=20$ \\
\hline

NSL-KDD & 0.942$\pm$0.000 & 0.955$\pm$0.006 & 0.954$\pm$0.007 &0.954$\pm$0.007& 0.955$\pm$0.007 &0.954$\pm$0.007& 0.953$\pm$0.006& \textbf{0.966}$\pm$0.000 & \textbf{0.966}$\pm$0.000\\
Spambase & 0.820$\pm$0.015 & 0.858$\pm$0.018 & 0.867$\pm$0.012 &0.911$\pm$0.016& \textbf{0.921}$\pm$0.016 &0.907$\pm$0.013& 0.867$\pm$0.012 & 0.862$\pm$0.013 & 0.859$\pm$0.011\\
Arrhythmia & 0.669$\pm$0.130 & 0.756$\pm$0.046 & 0.781$\pm$0.054 &0.789$\pm$0.053& \textbf{0.797}$\pm$0.048 &0.785$\pm$0.042& 0.780$\pm$0.044 & 0.780$\pm$0.036 & 0.784$\pm$0.044\\
Cardio & 0.976$\pm$0.010 & \textbf{0.999}$\pm$0.000 & \textbf{0.999}$\pm$0.000 &\textbf{0.999}$\pm$0.001& \textbf{0.999}$\pm$0.001 &\textbf{0.999}$\pm$0.000& \textbf{0.999}$\pm$0.001 & \textbf{0.999}$\pm$0.001 & 0.998$\pm$0.000 \\
Shuttle & 0.981$\pm$0.007 & 0.983$\pm$0.006 & \textbf{0.986}$\pm$0.004 &\textbf{0.986}$\pm$0.005& 0.983$\pm$0.004 &0.982$\pm$0.004& 0.982$\pm$0.005 &  0.980$\pm$0.005 & 0.974$\pm$0.002 \\
Satellite & 0.878$\pm$0.009 & 0.890$\pm$0.007 & 0.885$\pm$0.006 &0.902$\pm$0.005& \textbf{0.918}$\pm$0.007 & 0.917$\pm$0.005& 0.905$\pm$0.005 & 0.911$\pm$0.004 & 0.895$\pm$0.008\\
Mammography & 0.863$\pm$0.057 & 0.897$\pm$0.036 & 0.922$\pm$0.035 &0.958$\pm$0.048& \textbf{0.972}$\pm$0.002 &0.966$\pm$0.060& 0.941$\pm$0.042 & 0.941$\pm$0.046 & 0.906$\pm$0.041 \\
Fraud & 0.967$\pm$0.004 & 0.970$\pm$0.005 & 0.972$\pm$0.035 &0.977$\pm$0.006& \textbf{0.982}$\pm$0.003 &0.976$\pm$0.003& 0.965$\pm$0.003 & 0.962$\pm$0.007 & 0.956$\pm$0.010\\
\hline
Average & 0.887$\pm$0.029 & 0.914$\pm$0.016 &0.921$\pm$0.019 & 0.935$\pm$0.018 &  \textbf{0.941}$\pm$0.011 & 0.936$\pm$0.017 & 0.924$\pm$0.015 & 0.925$\pm$0.014 & 0.917$\pm$0.015 \\
\hline
\hline
\textbf{\multirow{2}{*}
{Dataset}}& \multicolumn{9}{c||}{\textbf{AUC-PR Performance}}  \\
\cline{2-10}  & $a_0=0.1$ & $a_0=1$ & $a_0=3$ &$a_0=4$& $a_0=5$ &$a_0=6$& $a_0=7$ & $a_0=10$  & $a_0=20$\\
\hline
NSL-KDD & 0.966$\pm$0.000 & 0.971$\pm$0.003 & 0.970$\pm$0.004 &0.969$\pm$0.004& 0.970$\pm$0.003 &0.968$\pm$0.003& 0.968$\pm$0.003 &\textbf{0.978}$\pm$0.000 &\textbf{0.978}$\pm$0.000 \\
Spambase & 0.835$\pm$0.014 & 0.872$\pm$0.012 & 0.883$\pm$0.006 &0.897$\pm$0.009& \textbf{0.910}$\pm$0.036 &0.901$\pm$0.008& 0.885$\pm$0.008 & 0.882$\pm$0.009 & 0.880$\pm$0.007  \\
Arrhythmia  & 0.483$\pm$0.136 & 0.644$\pm$0.049 & 0.673$\pm$0.046 &0.679$\pm$0.042& 0.681$\pm$0.038 &0.680$\pm$0.038& 0.678$\pm$0.040 &\textbf{0.689}$\pm$0.045 &0.680$\pm$0.041\\
Cardio & 0.959$\pm$0.011 & \textbf{0.991}$\pm$0.002 & 0.990$\pm$0.003 &0.990$\pm$0.004& 0.990$\pm$0.003 &0.989$\pm$0.003& 0.988$\pm$0.004 &0.989$\pm$0.004 & 0.987$\pm$0.003 \\
Shuttle & 0.970$\pm$0.007 & \textbf{0.976}$\pm$0.003 & 0.975$\pm$0.002 &0.975$\pm$0.002& 0.975$\pm$0.001 &0.974$\pm$0.001& 0.974$\pm$0.001 &0.973$\pm$0.002 & 0.970$\pm$0.002 \\
Satellite & 0.847$\pm$0.009 & 0.856$\pm$0.008 & 0.853$\pm$0.007 &0.882$\pm$0.006& \textbf{0.895}$\pm$0.006 &0.887$\pm$0.004& 0.889$\pm$0.004 &0.888$\pm$0.003 & 0.886$\pm$0.005 \\
Mammography & 0.498$\pm$0.027 & 0.485$\pm$0.051 & 0.505$\pm$0.052 &0.631$\pm$0.068& \textbf{0.645}$\pm$0.021 &0.610$\pm$0.064& 0.577$\pm$0.135 &0.598$\pm$0.134 & 0.573$\pm$0.139\\
Fraud & 0.652$\pm$0.034 & 0.689$\pm$0.021 & 0.692$\pm$0.009 &\textbf{0.693}$\pm$0.009& 0.692$\pm$0.006  &0.685$\pm$0.008& 0.684$\pm$0.008 &0.683$\pm$0.009 & 0.682$\pm$0.008 \\
\hline
Average & 0.776$\pm$0.029 & 0.811$\pm$0.019 & 0.818$\pm$0.016 & 0.840$\pm$0.018 & \textbf{0.845}$\pm$0.014 & 0.837$\pm$0.016 & 0.830$\pm$0.025 & 0.835$\pm$0.026 & 0.830$\pm$0.026  \\
\hline
\end{tabular}}
\end{spacing}
\label{tablefora0}
\end{table*}

\subsubsection{Effect of Hyper-parameters on Objective Function}

Next, we evaluate the effectiveness of the hyper-parameters in the objective function. We conduct two groups of experiments to investigate the impact of the hyper-parameter $a_0$ and $\lambda$, respectively. In the first group of experiments, we set $a_0$ to different values, \emph{i.e.}, 0.1, 1, 3, 4, 5, 6, 7, 10 and 20. In the second group of experiments, we set $\lambda$ to 0.1, 0.5, 1, 5 and 10, respectively. The other experimental settings are exactly the same as those of the proposed method described in Section IV-C.


As shown in Table \ref{tablefora0}, the performance of our proposed model tends to be better when $a_0$ is close to a specific value. On the one hand, if $a_0$ is too small, it is difficult to ensure that the model is discriminative for distinguishing anomalies from the normal ones. On the other hand , if $a_0$ is too large, the performance of the proposed model may drop due to the difficulties of training such an objective function. The experiment results also indicate that the hyper-parameter $a_0$ could be set to a fixed value (5  in  our experiments) in the proposed model. Moreover, We observe that the optimal choice of $a_0$ is consistent in all datasets. In Table \ref{tableforlambda}, the experiment results illustrate that the proposed method achieves the best performance consistently when the contributions of two parts to the entire loss are balanced by setting $\lambda$ to 1. This indicates the optimal hyper-parameter choice is stable across datasets. Also, we can seen that the performance corresponding to different $\lambda$ does not vary significantly (within 2.6\% on average). The above experimental results demonstrate that the setting of hyper-parameters in the proposed model is appropriate.



\subsubsection{Effect of Pre-training}

We then evaluate the effectiveness of the pre-training procedure adopted in our proposed model. We compare our method with a model that does not include the pre-training procedure (the first procedure described in Algorithm 1) in the training process. Note that the network structure and other experimental settings for the end-to-end training model are exactly the same as those of the proposed model. In contrast to the proposed method that loads the pre-trained weight parameters into the network before the end-to-end training of the whole network, the alternative model directly trains the whole network in an end-to-end manner.

As shown in Table \ref{tablepretrain}, compared to the model without a pre-training procedure, our proposed model achieves a significant improvement by 3.1\% (average) and 4.6\% (average) 

\begin{table*}[!ht]\scriptsize
\caption{\centering Experimental results (Mean $\pm$ standard deviation) to investigate on the effect of the hyper-parameter $\lambda$. }
\centering
\begin{spacing}{1.15}
\setlength{\tabcolsep}{1.8mm}{
\begin{tabular}{||c|ccccc||}
\hline
\textbf{\multirow{2}{*}
{Dataset}}& \multicolumn{5}{c||}{\textbf{AUC-ROC Performance}}  \\
\cline{2-6} & $\lambda=0.1$ & $\lambda=0.5$ & $\lambda=1$  & $\lambda=5$  & $\lambda=10$ \\
\hline

NSL-KDD & 0.941$\pm$0.014 & 0.937$\pm$0.013 & \textbf{0.955}$\pm$0.007 & 0.944$\pm$0.013 & 0.940$\pm$0.014\\
Spambase & 0.817$\pm$0.025 & 0.832$\pm$0.014 & \textbf{0.921}$\pm$0.016 & 0.901$\pm$0.012 & 0.814$\pm$0.026\\
Arrhythmia  & 0.783$\pm$0.051 & 0.796$\pm$0.051 & \textbf{0.797}$\pm$0.048 & 0.785$\pm$0.055 & 0.792$\pm$0.062\\
Cardio & \textbf{0.999}$\pm$0.000 & \textbf{0.999}$\pm$0.000 & \textbf{0.999}$\pm$0.001 & \textbf{0.999}$\pm$0.000 & \textbf{0.999}$\pm$0.000 \\
Shuttle & 0.977$\pm$0.009 & 0.974$\pm$0.008 & \textbf{0.983}$\pm$0.004 & 0.979$\pm$0.009 & 0.976$\pm$0.010\\
Satellite & 0.877$\pm$0.011 & 0.880$\pm$0.012 & \textbf{0.918}$\pm$0.007 & 0.897$\pm$0.011 & 0.878$\pm$0.010\\
Mammography & 0.968$\pm$0.005 & 0.967$\pm$0.002 & \textbf{0.972}$\pm$0.002 & 0.968$\pm$0.005 & 0.969$\pm$0.004 \\
Fraud & 0.968$\pm$0.006 & 0.967$\pm$0.008 & \textbf{0.982}$\pm$0.003  & 0.971$\pm$0.004 & 0.968$\pm$0.005 \\
\hline
Average & 0.916$\pm$0.015 & 0.919$\pm$0.014 & \textbf{0.941}$\pm$0.011  & 0.931$\pm$0.014 & 0.917$\pm$0.016 \\
\hline
\hline
\textbf{\multirow{2}{*}
{Dataset}}& \multicolumn{5}{c||}{\textbf{AUC-PR Performance}}  \\
\cline{2-6}  & $\lambda=0.1$ & $\lambda=0.5$ & $\lambda=1$  & $\lambda=5$  & $\lambda=10$ \\
\hline
NSL-KDD &0.964$\pm$0.007 & 0.963$\pm$0.007 & \textbf{0.970}$\pm$0.003 & 0.964$\pm$0.006 & 0.964$\pm$0.007 \\
Spambase & 0.840$\pm$0.016 & 0.853$\pm$0.011 & \textbf{0.910}$\pm$0.036 & 0.900$\pm$0.006 & 0.838$\pm$0.017  \\
Arrhythmia  & 0.648$\pm$0.037 & 0.670$\pm$0.038 & \textbf{0.681}$\pm$0.038 & 0.650$\pm$0.032 &0.648$\pm$0.046\\
Cardio & \textbf{0.990}$\pm$0.001 & \textbf{0.990}$\pm$0.002 & \textbf{0.990}$\pm$0.003 & \textbf{0.990}$\pm$0.001 & \textbf{0.990}$\pm$0.001 \\
Shuttle & 0.963$\pm$0.006 & 0.960$\pm$0.004 & \textbf{0.975}$\pm$0.001 & 0.973$\pm$0.006 & 0.963$\pm$0.006 \\
Satellite & 0.853$\pm$0.008 & 0.857$\pm$0.011 & \textbf{0.895}$\pm$0.006 & 0.863$\pm$0.009 & 0.854$\pm$0.008 \\
Mammography & 0.616$\pm$0.017 & 0.606$\pm$0.018 & \textbf{0.645}$\pm$0.021 & 0.610$\pm$0.018 & 0.618$\pm$0.019\\
Fraud & 0.680$\pm$0.014 & 0.679$\pm$0.014 & \textbf{0.692}$\pm$0.006  & 0.681$\pm$0.014 & 0.680$\pm$0.014 \\
\hline
Average & 0.819$\pm$0.013 & 0.822$\pm$0.013 & \textbf{0.845}$\pm$0.014  & 0.829$\pm$0.012 & 0.819$\pm$0.015  \\
\hline
\end{tabular}}
\end{spacing}
\label{tableforlambda}
\end{table*}

\noindent in AUC-ROC and AUC-PR, respectively. These results indicate that the pre-training procedure is an essential component of our proposed method. We postulate that pre-training will help the network avoid some bad local-minima and lead to a better local solution. 

\subsubsection{Effect of Feeding Reconstruction Error to Each Layer of Anomaly Score Generator}

In the proposed method, we feed the reconstruction error $e$ to each layer of the MLP in the anomaly score generator. To investigate the effect of using this design, we construct a variant which does not feed error $e$ to each layer but just concatenate it with the other two factors at the first layer. 

The result is shown in Table \ref{tableextra}. Comparing with the model using the original design, the model employing this variant strategy has led to an average performance drop of 1.5\% and 1.4\% on AUC-ROC and AUC-PR, respectively. This experimental result demonstrates that it is necessary to use the proposed design to avoid losing information of the reconstruction error. 

\begin{table}[!h]\scriptsize
\caption{\centering Experimental results (Mean $\pm$ standard deviation) to investigate the Effect of the pre-training procedure. }
\centering
\begin{spacing}{1.15}
\setlength{\tabcolsep}{1.5mm}{
\begin{tabular}{||c|cc|cc||}
\hline
\textbf{\multirow{2}{*}
{Dataset}}& \multicolumn{2}{c|}{\textbf{AUC-ROC Performance}} &\multicolumn{2}{c||}{\textbf{AUC-PR Performance}}  \\
\cline{2-3} \cline{4-5} & \textbf{Pre-trained} & \textbf{End-to-End} & \textbf{Pre-trained} & \textbf{End-to-End}\\
\hline
NSL-KDD & \textbf{0.955}$\pm$0.007 & 0.953$\pm$0.000 & \textbf{0.970}$\pm$0.003 & 0.966$\pm$0.000  \\
Spambase & \textbf{0.921}$\pm$0.016 & 0.771$\pm$0.016 & \textbf{0.910}$\pm$0.036 & 0.805$\pm$0.013 \\
Arrhythmia  & \textbf{0.797}$\pm$0.048 & 0.783$\pm$0.056 & \textbf{0.681}$\pm$0.038 & 0.648$\pm$0.039  \\
Cardio & \textbf{0.999}$\pm$0.001 & 0.991$\pm$0.010 & \textbf{0.990}$\pm$0.003 & 0.964$\pm$0.018  \\
Shuttle & \textbf{0.983}$\pm$0.004 & 0.976$\pm$0.008 & \textbf{0.975}$\pm$0.001 & 0.963$\pm$0.005 \\
Satellite & \textbf{0.918}$\pm$0.007 & 0.877$\pm$0.011 & \textbf{0.895}$\pm$0.006 & 0.854$\pm$0.009  \\
Mammography & \textbf{0.972}$\pm$0.002 & 0.953$\pm$0.001 & \textbf{0.645}$\pm$0.021 & 0.634$\pm$0.011  \\
Fraud & \textbf{0.982}$\pm$0.003 & 0.968$\pm$0.005 & \textbf{0.692}$\pm$0.006&  0.682$\pm$0.013  \\
\hline
Average & \textbf{0.941}$\pm$0.011 & 0.909$\pm$0.013 & \textbf{0.845}$\pm$0.014  & 0.815$\pm$0.014  \\
\hline
\end{tabular}}
\end{spacing}
\label{tablepretrain}
\end{table}

\begin{table}[!h]\scriptsize
\caption{\centering Experimental results (Mean $\pm$ standard deviation) to investigate the Effect of Feeding Reconstruction Error to Each Layer of Anomaly Score Generator. }
\centering
\begin{spacing}{1.15}
\setlength{\tabcolsep}{1.5mm}{
\begin{tabular}{||c|cc|cc||}
\hline
\textbf{\multirow{2}{*}
{Dataset}}& \multicolumn{2}{c|}{\textbf{AUC-ROC Performance}} &\multicolumn{2}{c||}{\textbf{AUC-PR Performance}}  \\
\cline{2-3} \cline{4-5} & \textbf{Proposed} & \textbf{Variant} & \textbf{Proposed} & \textbf{Variant}\\
\hline
NSL-KDD & \textbf{0.955}$\pm$0.007 & 0.949$\pm$0.005 & \textbf{0.970}$\pm$0.003 & 0.967$\pm$0.003  \\
Spambase & \textbf{0.921}$\pm$0.016 & 0.872$\pm$0.012 & \textbf{0.910}$\pm$0.036 & 0.887$\pm$0.007 \\
Arrhythmia  & \textbf{0.797}$\pm$0.048 & 0.792$\pm$0.034 & \textbf{0.681}$\pm$0.038 & 0.679$\pm$0.025  \\
Cardio & \textbf{0.999}$\pm$0.001 & \textbf{0.999}$\pm$0.000 & \textbf{0.990}$\pm$0.003 & \textbf{0.990}$\pm$0.003  \\
Shuttle & \textbf{0.983}$\pm$0.004 & 0.982$\pm$0.004 & \textbf{0.975}$\pm$0.001 & \textbf{0.975}$\pm$0.001 \\
Satellite & \textbf{0.918}$\pm$0.007 & 0.883$\pm$0.006 & \textbf{0.895}$\pm$0.006 & 0.856$\pm$0.007  \\
Mammography & \textbf{0.972}$\pm$0.002 & 0.896$\pm$0.041 & \textbf{0.645}$\pm$0.021 & 0.540$\pm$0.064  \\
Fraud & \textbf{0.982}$\pm$0.003 & 0.968$\pm$0.004 & \textbf{0.692}$\pm$0.006&  0.681$\pm$0.014  \\
\hline
Average & \textbf{0.941}$\pm$0.011 & 0.918$\pm$0.013 & \textbf{0.845}$\pm$0.014  & 0.822$\pm$0.016  \\
\hline
\end{tabular}}
\end{spacing}
\label{tableextra}
\end{table}

\section{Conclusion}
In this paper, we address the weakly-supervised anomaly detection problem that there are only a few labeled abnormal data and sufficient normal data available for training. We propose a novel encoding strategy that transforms the input data into a more meaningful representation, in which typical anomalies could be significantly deviated from the normal ones. We then propose an anomaly detection model that seamlessly incorporates the information from the proposed encoding. To cope with the limitation of data insufficiency for annotated anomalies, the proposed model consists of two parts, \emph{i.e.}, the feature encoder and the anomaly score generator, which are jointly optimized through the training process with a specially designed deviation loss. Extensive experiments demonstrate that the proposed method could effectively leverage the limited-but-useful information from the few labeled abnormal data and achieve superior performance over the competitive methods. In the future work, we plan to explore multi-task learning\cite{zeng2018multi} and nonlinear correlation analysis\cite{qin2016demographic},\cite{liu2020learning},\cite{chen2020citywide} to further utilize the implicit knowledge from available data.


%



\section*{Acknowledgment}
We would like to thank Yiran Deng at Sichuan University for proofreading this paper.

\ifCLASSOPTIONcaptionsoff
  \newpage
\fi



\bibliographystyle{IEEEtran}
\bibliography{bare_jrnl.bbl}
\end{document}